# An Efficient Image Retrieval Based on Fusion of Low-Level Visual Features

**Atif Nazir[1](Non-member) , Kashif Nazir[2] (Non-Member)**

[1] Department of Computer Science, National Textile University, Faisalabad, Pakistan.
[atif.nazir1225@gmail.com]

[2] Department of Computer Science, Riphah International University , Faisalabad, Pakistan.
[kashif@riphahfsd.edu.pk]

*Corresponding author: **Atif Nazir**

## *Abstract*

Due to an increase in the number of image achieves, Content-Based Image Retrieval (CBIR) has gained attention for research community of computer vision. The image visual contents are represented in a feature space in the form of numerical values that is considered as a feature vector of image. Images belonging to different classes may contain the common visuals and shapes that can result in the closeness of computed feature space of two different images belonging to separate classes. Due to this reason, feature extraction and image representation is selected with appropriate features as it directly affects the performance of image retrieval system. The commonly used visual features are image spatial layout, color, texture and shape. Image feature space is combined to achieve the discriminating ability that is not possible to achieve when the features are used separately. Due to this reason, in this paper, we aim to explore the low-level feature combination that are based on color and shape features. We selected color moments and color histogram to represent color while shape is represented by using invariant moments. We selected this combination, as these features are reported intuitive, compact and robust for image representation. We evaluated the performance of our proposed research by using the Corel, Coil and Ground Truth (GT) image datasets. We evaluated the proposed low-level feature fusion by calculating the precision, recall and time required for feature extraction. The precision, recall and feature extraction values obtained from the proposed low-level feature fusion outperforms the existing research of CBIR.

***Keywords:*** Image retrieval, color features, shape features, low-level features combination,

# 1. Introduction

Due to recent growth in the volume of image contents, the search of relevant image from a database is an open research problem for computer vision community [1-3]. The use of internet and social media is growing day-by-day and on daily basis, there is sufficient addition of images to the existing archives [3, 4]. Due to these reasons, there is a need to improve the existing CBIR systems so that they can handle the transformations that are part of image archives [5, 6]. In past, the images were labeled manually and retrieved by using keyword-based search . The manual efforts and difference in human perception to interpret the images make this approach less effective and reliable for the existing image datasets [1, 2, 5]. The most of the existing search engines retrieve the images by using manual tagging that involves human efforts and traditional text-based approaches [1, 2, 5, 7].

The second approach to retrieve the images is based on the use of visual image features that are color, shape, texture and spatial image layout [2, 5, 6]. Query By Image Contents (QBIC), SIMPLicity and Blobworld are the examples of image search systems that are using low-level image features [2, 3]. The main aim of any CBIR research is to sort the images in order of closeness that are visually similar [7-10]. The feature extraction approaches are applied to generate the discriminative representation of images and this is considered as one of the key challenges in CBIR [2, 3]. The fusion of low-level features is used to boost the performance of image retrieval as single feature-based image representation cannot handle the variety of changes/transformations that are in the multimedia archives [7-10]. The similarity measure is applied in CBIR to sort the similar images and selection of appropriate similarity measure affects the performance of CBIR [5].

In last few years, assortment of low level visual features is proposed in the literature that are associated with different applications of CBIR [11] . The low-level visual features are divided into two main types: 1) global and 2) local. The global features are applied to extract the visual information that is in the form of color, shape, texture and spatial contents by representing the whole image [5, 6]. The local features describe the image parts such as corner, image edges or the regions obtained through image segmentation [5, 6]. The type of selected image representation (either global or local) is dependent on the user requirements [8]. The appropriate selection of two low-level features can increase the effectiveness of  CBIR because this process integrates different feature spaces[8]. Here it is necessary to mention that the proper selection of features to be use in fusion is important as fusion of inappropriate feature spaces degrades the performance of CBIR [5, 6, 8]. Due to these facts, in this paper, we aim to investigate the low-level feature fusion of color  and shape features ( color histogram [12] , color moments [13, 14] and invariant moments [15]) as these features are reported intuitive, compact and robust for image representation, rotations and translation [3, 16]. In this paper, we present a simple idea with very less computational cost that combines the low-level features for CBIR. Therefore, our main contributions are 1) design an efficient image retrieval framework by using a combination of pre-existed low-level visual features based on color and shape, 2) reduction of semantic gap. By using a combination of these three features, our proposed framework performs better than many of image retrieval frameworks requiring lot of computational power. It is worth emphasizing that we selected the features that are pre-existed and we just performed feature fusion, as we want to deliver an efficient way for image retrieval.

The remaining paper is organized as follows: **Section 2** presents literature review, **Section 3** presents the proposed research methodology, **Section 4** is about results and discussions while Section 5 concludes the proposed research.

## 2. Related Work

In last two decades, the research of CBIR is focused on different techniques that are based on computation of image spatial layout, features integration/fusion for hybrid image representation [8, 9, 11, 16]. Among low-level image features, color is used in image analysis and retrieval to represent objects and it contains the information that can differentiate between the foreground and background. Color features are unique as they are considered to be invariant to the transformation such as rotations [2, 3, 16]. According to Jiao et al. [17], the fusion of Color Scale Invariant Feature Transform and Edge Orientation Difference Histogram (EODH) can enhance the effectiveness of CBIR. Edge Orientation Difference Histogram (EODH) is selected for the image feature vector representation to achieve rotation and scale invariance. In another research, Walia et al. [18], investigated the fusion of low-level features that are based on texture and color representation. The two low-level features are selected to be used on a combination to obtained a better CBIR performance. Dubey et al. [19], investigated the another fusion of low-level features. The RGB color space is used in a quantized form and texture is computed by using different patterns. The final feature vector representation contains both of low-level features and the selected fusion enhances the CBIR performance. Ali et al. [8], investigated the visual word fusion of two local features. Two different codebooks are computed from SIFT and Speeded-Up Robust Features (SURF)and later on an image is represented in the form of clusters of two local features. The proposed techniques for feature vector representation is reported beneficial as it increases the size of image histogram by representing it with two distant features. In an other research [11], local features are integrated with binary feature space to represent an image as the histogram of SIFT and FREAK. The research studies [8, 11] show that the integration of two feature space is beneficial for image retrieval problem. Yuan et al. [20], selected a combination of local features and Local Binary Pattern (LBP) to address the problem that occur due to noise in the background when SIFT is used alone. The research of Yuan et al. [20] is based on early feature fusion balance is maintained by applying a weighted scheme at the time of computation of cluster. According to [21]. the high computational pre-trained Convolutional Neural Networks (CNN) models can be used for image retrieval tasks. The use of CNN along with with proper feature refining schemes outperforms hand-crafted features on all image datasets

The approaches of CBIR that are discussed earlier [7-11, 20, 22] are solving the problem of image retrieval by using image classification model that is based on training and testing approach and require lot of training time when training data is large. The research that is proposed in this paper is using a combination of low-level visual features that are based on color and shape. The computed feature space is used as image representation and without the use of any classification model, we performed the image retrieval through distance measures that is beneficial as it save the time that is associated with classifier training.

## 3. Proposed Methodology

The proposed research presented in this paper is based on the fusion of low-level image features that are based on color and shape. The detail about color histogram, color moments, invariant moments and feature fusion is mentioned in the following subsections.

### a. Color Histogram

Color histogram [12] is used to represent the color information about the image. A color image consists of red, green and blue planes and is referred as RGB image. The color histogram can be constructed in other color spaces such as HSV. The color bins are represented in the form of histograms and the total number of bins in RGB histograms can be determined on the basis of RGB planes.

## 3.2 Color Moments

Color moments [13, 14] are used to overcome the quantization impact of color histogram. For the color distribution, the color moments are divided into three central moments that are mean, skewness and standard deviation. The mathematical equations of mean, skewness and standard deviation are presented as [14]:

I. First-moment (Mean)

$$E_i = \sum_{N}^{j=1} \left| \frac{1}{N} P_{ij} \right| \qquad (1)$$

It calculates the average color distribution of an image. Here N is the number of pixels in the image and $P_{ij}$ is the j pixel at i channel

II. Second-moment (standard deviation)

$$\sigma_i = \sqrt{\left| \frac{1}{2} \sum_{N}^{j=1} (P_{ij} - E_i)^2 \right|} \qquad (2)$$

The standard deviation is calculated by the square ro ot of the variance for color distribution and *Ei* is the mean is calculated in equation 1.

III. Third-moment (skewness)

$$S_i = \sqrt[3]{\left| \frac{1}{N} \sum_{N}^{j=1} (P_{ij} - E_i)^3 \right|} \qquad (3)$$

Skewness *Si* is used to find the asymmetric in the color distribution is.
The sum of weighted difference and the image distribution between the two images is 25 give as:

$$d_{mom}(X,Y) = \sum_{n=1}^{r} w_{n1} |M_n^1 - M_n^2| + w_{n2} |\sigma^1 - \sigma^2| + w_{n3} |S_i^1 - S_i^2| \qquad (4)$$

Here X and Y are two images, n is current channel for HSV H=1, S=2 and V=3, *M* 1 , *M* 2 is the first moment of mean between the two images, $\sigma$ 1 $\sigma$ 2 is the second moments of standard deviation between the two images, *S* 1, *S* 2 is the third moments skewness between the two images and *w* that represents the assigned weight.

## 3.3 Invariant Moment

There are seven invariant moments for shape features [15] that are reported efficient incase of transformations such as change in view point or illuminations [23]. These moments are robust to translation, in-plane rotations and scale invariant and are derived from the second and third order moment. Suppose *I(p, q)* is denoted by an image, so the moment for the order *x + y* and calculation of *mx,y* is

$$\mu_{x,y} = \sum_p \sum_q |p^x q^y I(p,q)| \qquad (5)$$

The central moment of an image derived as:

$$m_{x,y} = \sum_p \sum_q (p - pc)^x (q - qc)^y I(p, q) \qquad (6)$$

where $pc = \frac{m_{1,0}}{m_{0,0}}$ and $qc = \frac{m_{0,1}}{m_{0,0}}$ is the central region of the image $I(p, q)$ and the central moment for third order is derived as

$$\mu_{0,0} = m_{0,0}$$
$$\mu_{1,0} = \mu_{1,0} = 0$$
$$\mu_{2,0} = m_{2,0} - pc * m_{1,0}$$
$$\mu_{0,2} = m_{0,2} - qc * m_{0,1}$$
$$\mu_{1,1} = m_{1,1} - pc * m_{1,0}$$
$$\mu_{2,1} = m_{2,1} - 2pc * m_{1,1} - qc * m_{0,2} + m_{1,0}\, pc^2$$
$$\mu_{1,2} = m_{1,2} - 2qc * m_{1,1} - pc * m_{0,2} + m_{1,0}\, qc^2$$
$$\mu_{3,0} = m_{3,0} - 3pc * m_{2,0} + 2m_{1,0}\, pc^2$$
$$\mu_{0,3} = m_{0,3} - 3qc * m_{0,2} + 2m_{1,0}\, qc^2$$

The central moment of normalized form is denoted as $N_{x,y}$

$$N_{x,y} = \frac{\mu_{x,y}}{\mu_a} \qquad (7)$$

Where $a = \frac{x+y+2}{2}$ For the second and third moments, in this way we can calculate the 50 seven invariant moments that are:

$$w_1 = N_{2,0} + N_{0,2}$$
$$w_2 = (N_{2,0} - N_{0,2})^2 + 4N_{1,1}^2$$
$$w_3 = (N_{3,0} - N_{1,2})^2 + (3N_{2,1} - N_{0,3})^2$$
$$w_4 = (N_{3,0} - N_{1,2})^2 + (N_{2,1} - N_{0,3})^2$$
$$w_5 = (N_{3,0} - 3N_{1,2}) + (N_{3,0} + N_{1,2})[N_{3,0} + N_{1,2}]^2 - 3[N_{2,1} + N_{0,3}]^2 + 3N_{2,1}$$
$$- N_{0,3}(N_{2,1} - N_{0,3})\, 3\left[(N_{3,0} - N_{1,2})^2 - (N_{2,1} - N_{0,3})^2\right]$$
$$w_6 = (N_{2,0} - N_{0,2}) + (N_{3,0} - N_{1,2})^2 - (N_{2,1} - N_{0,3})^2 + 4N_{1,1}(N_{0,3} + N_{1,2})(N_{2,1} + N_{0,3})$$
$$w_7 = 3(N_{3,1} - N_{0,3})(N_{3,0} - N_{1,2})\left[(N_{3,0} - N_{1,2})^2\right] - 3(N_{2,1} - N_{0,3})^2 + 3N_{2,1}$$
$$- N_{0,3}(N_{2,1} - N_{0,3})\left[3(N_{3,0} + N_{1,2})^2 - (N_{2,1} + N_{0,3})^2\right]$$

**3.4 Proposed Method Based on Fusion of Color and Shape Features**

The block diagram of proposed research based on feature fusion of color and shape features is represented in **Fig. 1**. The main steps of the proposed method are described as:
1. In the pre-processing step, the images are resized and we used a resolution of 384*256 for every image that is available in the dataset.
2. The RGB images are converted into HSV color space and quantization is applied to construct a normalized HSV color histogram consisting of 32*2*2 bins.
3. For color moments, the RGB color space is converted into YCbCr (Y is luma and Cb,Cr are the blue,red difference luma). The YCbCr color space is used to extract first two moments that are mean and standard deviation and two channels are represented with a dimension of 1*6.
4. The invariant moments are calculated by using the seven moments for shape feature and as a result the length of obtained feature vector is 1*7.

5. The computed color histogram, color moments and invariant moments are combined to represent an image with combination of these three features. The resultant size of the feature vector for the image representation consists of 141 dimensions.
6. For image retrieval process, the query image is taken and steps are repeated from 1-5 to compute a feature vector on the basis of combination of low-level features consisting of color and shape. The similarity among the images is computed through Manhattan distance to sort the similar images from the database.

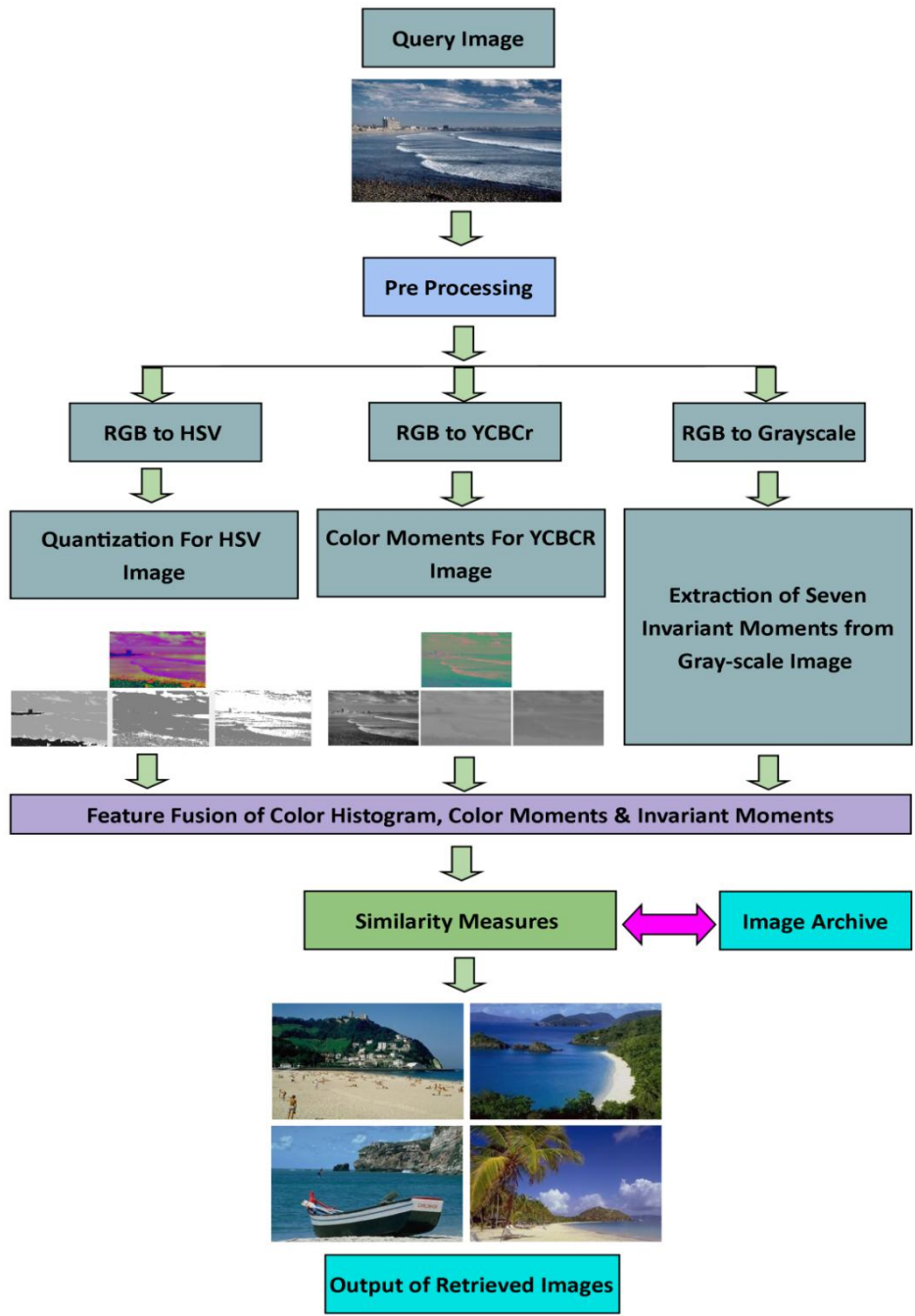

**Fig 1.** The block diagram of the proposed research.

# 4. Results and Discussion

We selected the Corel-1k [24] , Corel-1.5K [24], Coil [10] and Ground-Truth (GT) [8] image dataset to evaluate our proposed low-level feature fusion. These datasets are selected for performance evaluation as they are used in research published in recent years [1, 8, 17, 22, 25-27]. . We used MATLAB for the implementation of the proposed research and we selected precision [8] and recall [8] to compare the performance of our proposed model with the existing research of CBIR. To compare the efficiency, image retrieval time for feature extraction is compared with the recent research of CBIR [17, 19].

## 4.1 Image Retrieval Performance Using Corel-1k Image Dataset

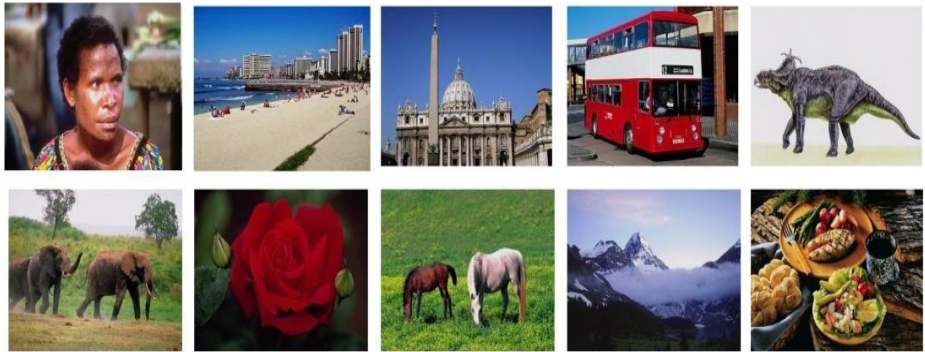

**Fig 2.** Images taken from each class of Corel-1K dataset [24].

The selected Corel-1K image dataset [24] contains one thousand images that are evenly divided into ten different classes (hundred images for each class). **Fig. 2** represents the gallery of images taken from each class of Corel-1K image dataset. represents the photo gallery of Corel-1K image dataset. We selected the numerical values for top-20 image retrievals and compared the average precision and recall values with the existing published research [8, 17, 22, 25], as the existing work is also reported for the selected value (top-20 retrievals). We also evaluated the performance of color histogram, color moments and invariant moments separately so that the importance of the proposed feature fusion can be elaborated. The category-wise results and comparisons for image retrieval for color histogram, color moments, invariant moments and proposed feature fusion is represented in **Table 1**. **Table 2-3** present numerical values comparison of precision and recall of proposed low-level feature fusion with existing research.

**Table 1.** Comparison of precision using invariant moments, color histogram, color moments and proposed method.

| Class name and method | Invariant Moment | Color Moments | Color histogram | **Proposed Research** |
|---|---|---|---|---|
| **Africa** | 25 | 75 | 65 | 75 |
| **Beach** | 15 | 60 | 35 | 65 |
| **Buildings** | 25 | 75 | 45 | 85 |
| **Busses** | 45 | 100 | 75 | 100 |
| **Dinosaurs** | 95 | 100 | 95 | 100 |
| **Elephants** | 25 | 75 | 35 | 85 |
| **Flowers** | 65 | 80 | 70 | 95 |
| **Horses** | 45 | 100 | 75 | 100 |
| **Mountains** | 20 | 35 | 75 | 55 |
| **Food** | 45 | 25 | 35 | 55 |
| **Mean** | 40.5 | 62.6 | 60.5 | 81.5 |

**Table 2**. Comparison of precision of the proposed low-level feature fusion with existing research.

| Class name and method | Visual words fusion [8] | Color –SIFT [17] | Spatial BoF [25] | SIFT-LBP [22] | **Proposed Research** |
|---|---|---|---|---|---|
| **Africa** | 60.08 | 74.6 | 64 | 57 | 75 |
| **Beach** | 60.39 | 37.8 | 54 | 58 | 65 |
| **Buildings** | 69.66 | 53.9 | 53 | 43 | 85 |
| **Busses** | 93.65 | 76.7 | 94 | 93 | 100 |
| **Dinosaurs** | 99.88 | 99 | 98 | 98 | 100 |
| **Elephants** | 70.76 | 65.9 | 78 | 58 | 85 |
| **Flowers** | 88.37 | 91.2 | 71 | 83 | 95 |
| **Horses** | 82.77 | 86.9 | 93 | 68 | 100 |
| **Mountains** | 61.08 | 58.5 | 42 | 46 | 55 |
| **Food** | 65.09 | 62.2 | 50 | 53 | 55 |
| **Mean** | 75.17 | 72.67 | 69.7 | 65.7 | 81.5 |

**Table 2.** Comparison of recall of the proposed low-level feature fusion with existing research.

| Class name and method | Visual words fusion [8] | Color –SIFT [17] | Spatial BoF [25] | SIFT-LBP [22] | **Proposed Research** |
|---|---|---|---|---|---|
| **Africa** | 12.02 | 14.92 | 12.8 | 11.4 | 15.0 |
| **Beach** | 12.08 | 7.56 | 10.8 | 11.6 | 13.0 |
| **Buildings** | 13.93 | 10.78 | 10.6 | 8.6 | 17.0 |
| **Busses** | 18.73 | 19.34 | 18.8 | 18.6 | 20.0 |
| **Dinosaurs** | 19.98 | 19.8 | 19.6 | 19.6 | 20.0 |
| **Elephants** | 14.15 | 13.18 | 15.6 | 11.6 | 17.0 |
| **Flowers** | 17.67 | 18.24 | 14.2 | 16.6 | 19.0 |
| **Horses** | 16.55 | 17.38 | 18.6 | 13.6 | 20.0 |
| **Mountains** | 12.22 | 11.7 | 8.4 | 9.2 | 11.0 |
| **Food** | 13.02 | 12.44 | 10.0 | 10.6 | 11.0 |
| **Mean** | 15.03 | 14.53 | 13.94 | 13.14 | 16.3 |

The numerical values of precision presented in **Table 1** for top-20 retrievals present the effectiveness of proposed research. The comparison of precision and recall presented in **Table 2-3** show a remarkable performance of the proposed research in most of the categories/classes as compared to the existing research [8, 17, 22, 25]. The mean image retrieval performance of proposed research is higher than that of color histogram, color moments and invariant moments. **Table 4** represents the time required for feature extraction and numbers of images retrieved in seconds. The numerical values presented in **Table 4** validate the robust performance of the proposed approach in term of feature extraction. Image retrieval results using different classes are represented in **Fig. 3-6.**

Table 4. **Time required for feature extraction.**

| Proposed Approach | RSHD [19] | Color-SIFT [17] | CDH [19] |
|---|---|---|---|
| **0.32** | 0.375 | 5.6 | **1.7** |

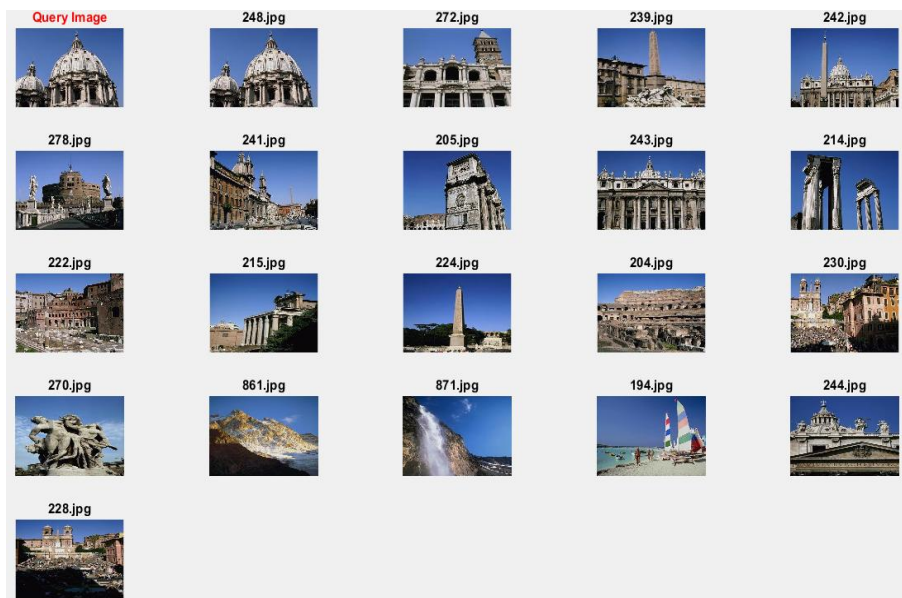

**Fig 3.** Query image result for the category "Building".

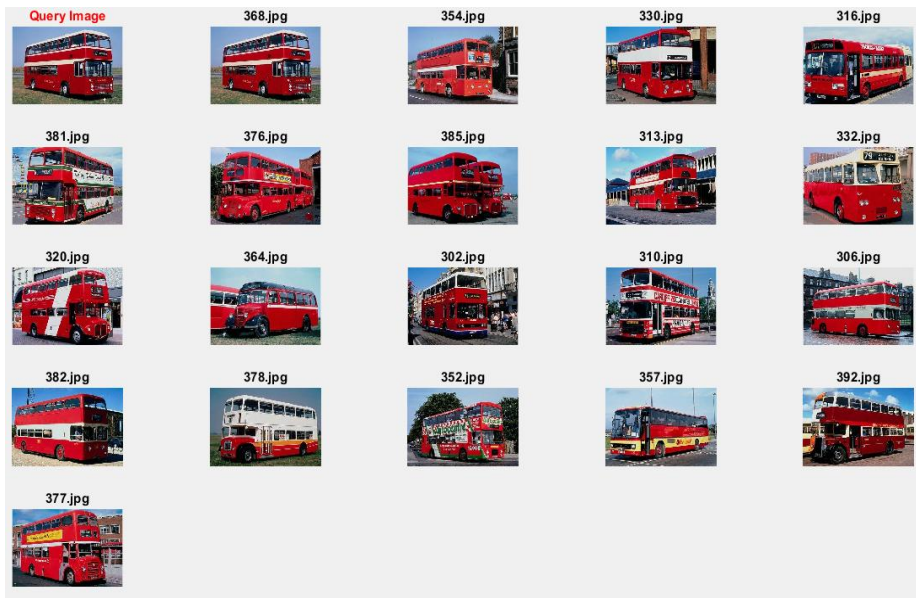

**Fig 4.** Query image result for the category "Busses".

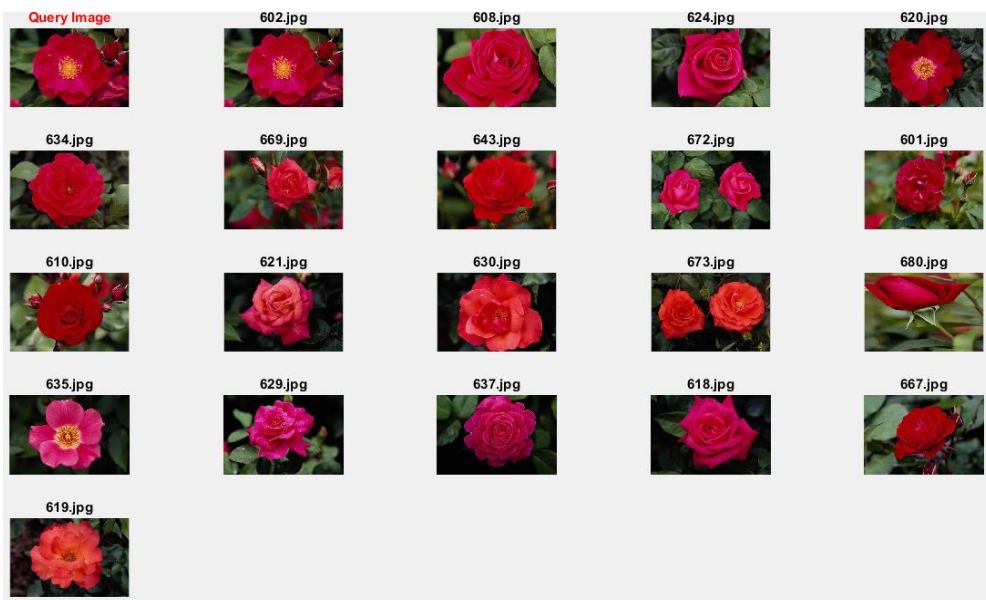

Fig 5. Query image result for the category " Flowers".

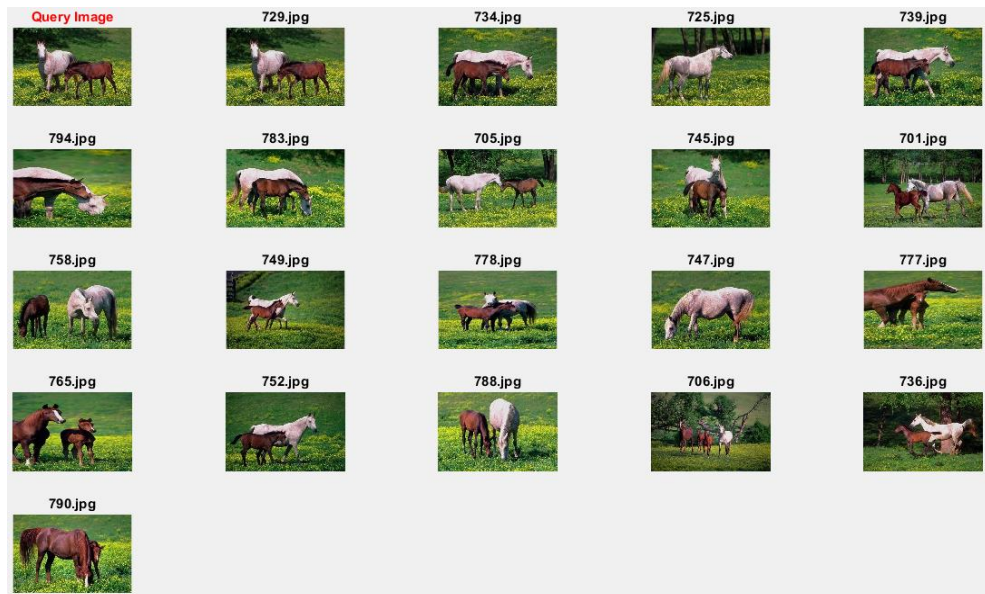

Fig 6. Query image result for the category "Horses".

## 4.2 Image Retrieval Performance Using Corel-1.5k Image Dataset

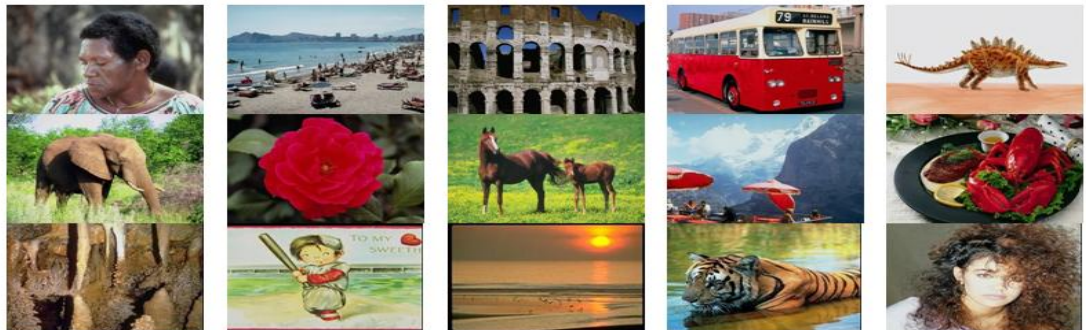

Fig 7. Images taken from each class of Corel-1.5 K dataset [24].

Corel-1.5K dataset consists of 15 classes and is previously used to evaluate CBIR research. **Fig. 7** represents the photo gallery of each class from Corel-1.5K image dataset. The details about the total number of classes and images in the Corel 1.5K dataset is available in [8, 28]. The average precision and recall values comparison of the proposed research are presented in **table. 5.**

Table 5. Average precision and recall values comparison obtained using the Corel 1.5K dataset.

| Performance / Method | Visual word fusion [[8] | SQ+ Spatiogram [28] | SQ+ mSpatiogram [28] | Proposed research |
|---|---|---|---|---|
| **Precision** | 74.95 | 63.95 | 74.1 | **80.5** |
| **Recall** | 14.99 | 12.79 | 14.82 | **16.1** |

According to the numerical values presented in table 5, the mean precision value obtained from the proposed research using Corel 1.5K dataset is higher than the recent research [8, 28]. Image retrieval results when using different classes of query images are shown in Fig. 8-9.

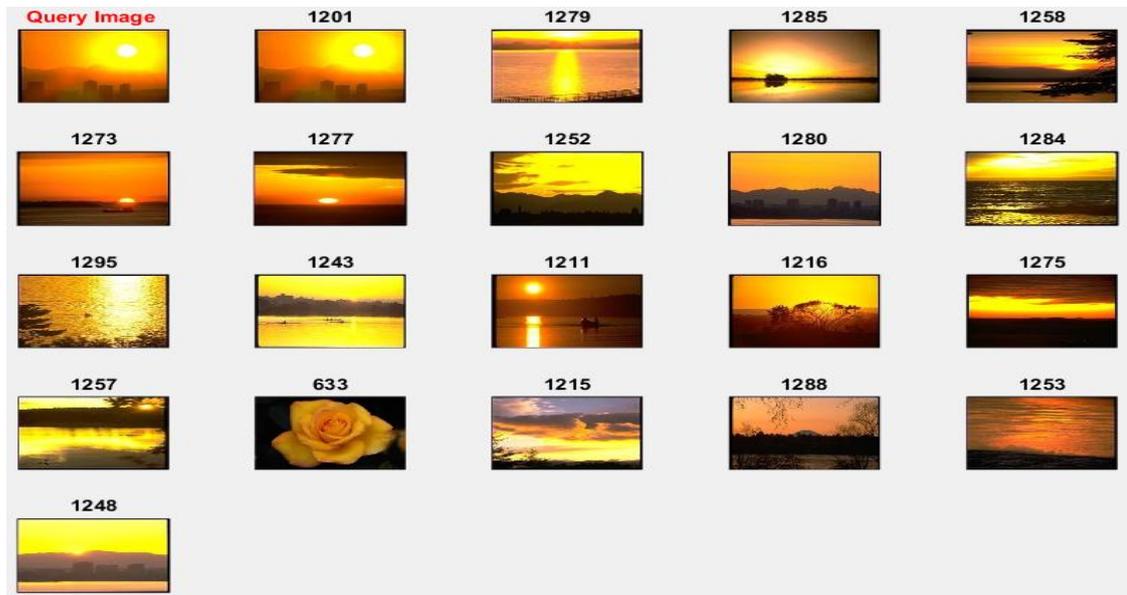

**Fig 8.** Query image result for the category " Sunset".

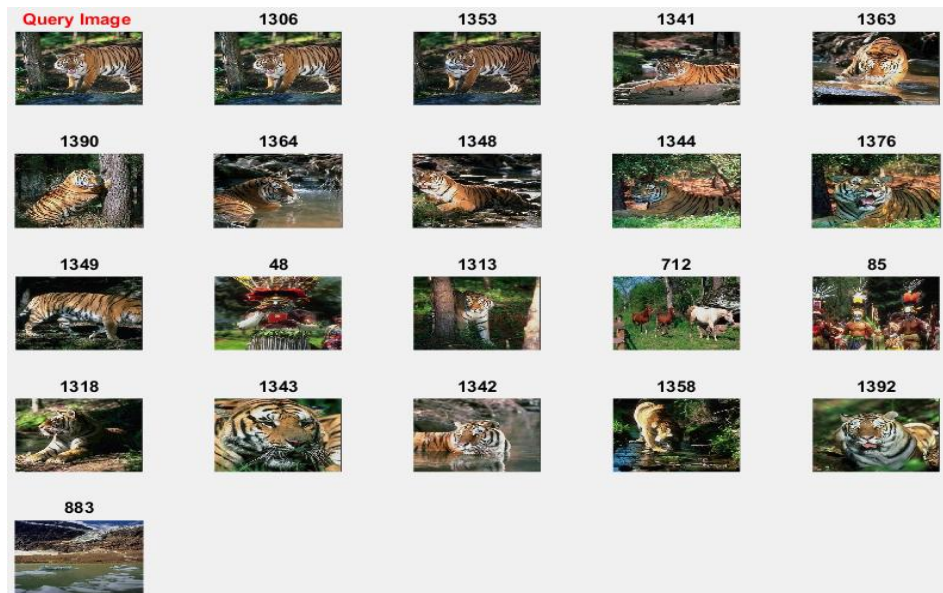

**Fig 9**. Query image result for the category "Tiger".

### 4.3 Image Retrieval Performance Using Coil Image Dataset

The Coil image dataset is publicly available and is used to evaluate research based on CBIR [29]. The comparison of precision and recall is presented graphically in **Fig. 10**. On the basis of experimental results, we can say that the proposed system is more reliable and outperforms the existing research [29]. Image retrieval results obtained using Coil image dataset are represented in **Fig. 11** .

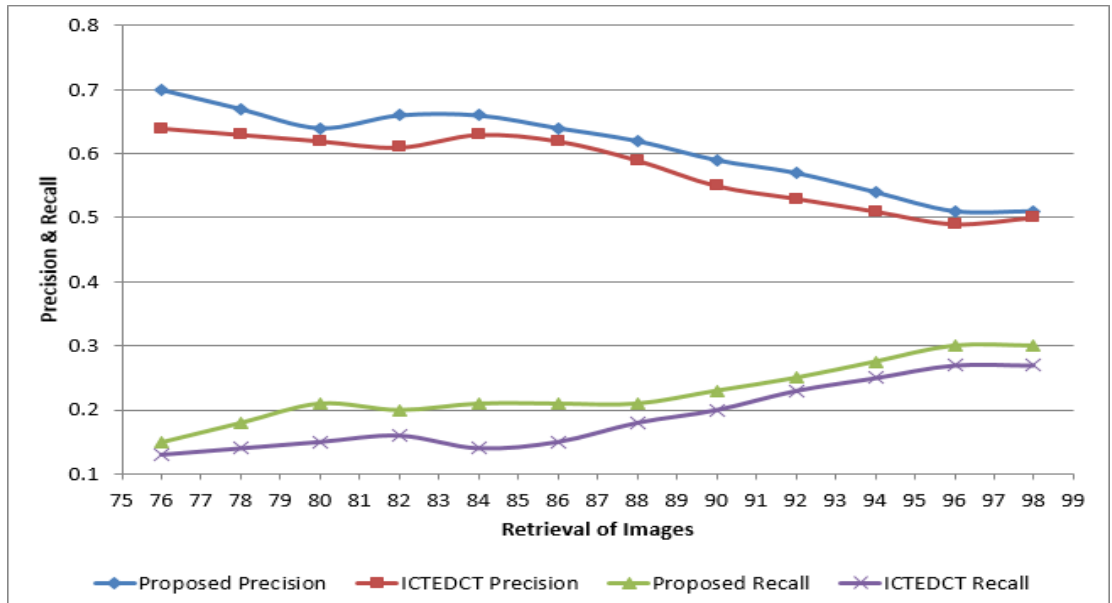

**Fig10**. Comparison of precision and recall using Coil dataset.

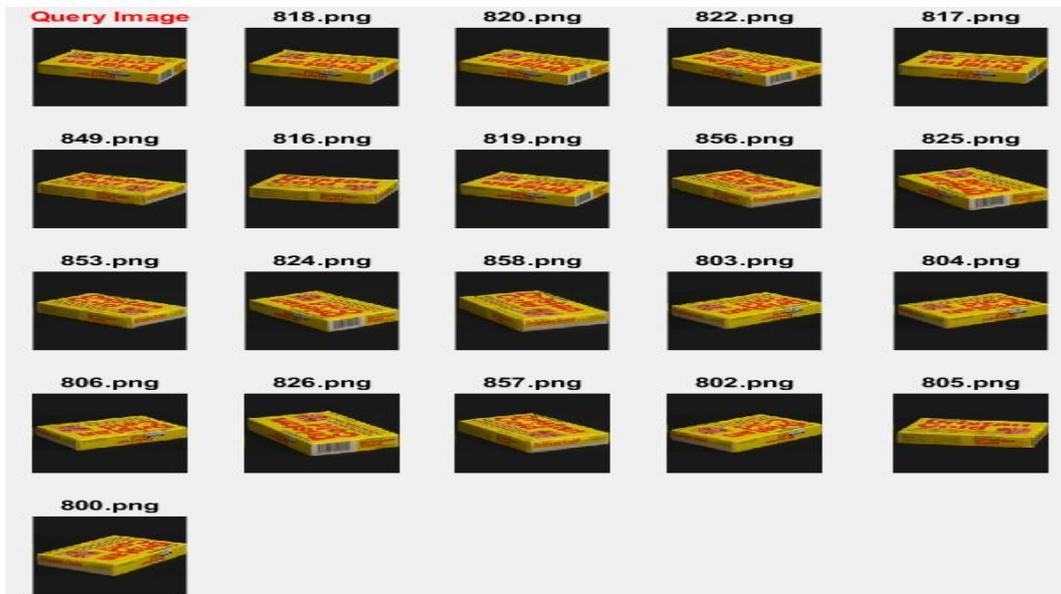

**Fig 11.** Query image result for Coil dataset.

## 4.4 Image Retrieval Performance Using GT Dataset

GT dataset is publicly available and is used to evaluate the research based on CBIR [8, 27, 30]. The details about the total number of classes and images in the GT dataset is available in [8, 27, 30]. **Table. 6** presents the comparison of mean precision values obtained using GT dataset.

Table. 6 Mean precision when using GT dataset.

| Performance / Method name | Visual words integration **[8]** | SVM-ensambles [30] | Wavelet- based IR [27] | **Proposed research** |
|---|---|---|---|---|
| Mean | 83.53 | 81.33 | 62.8 | 85.5 |

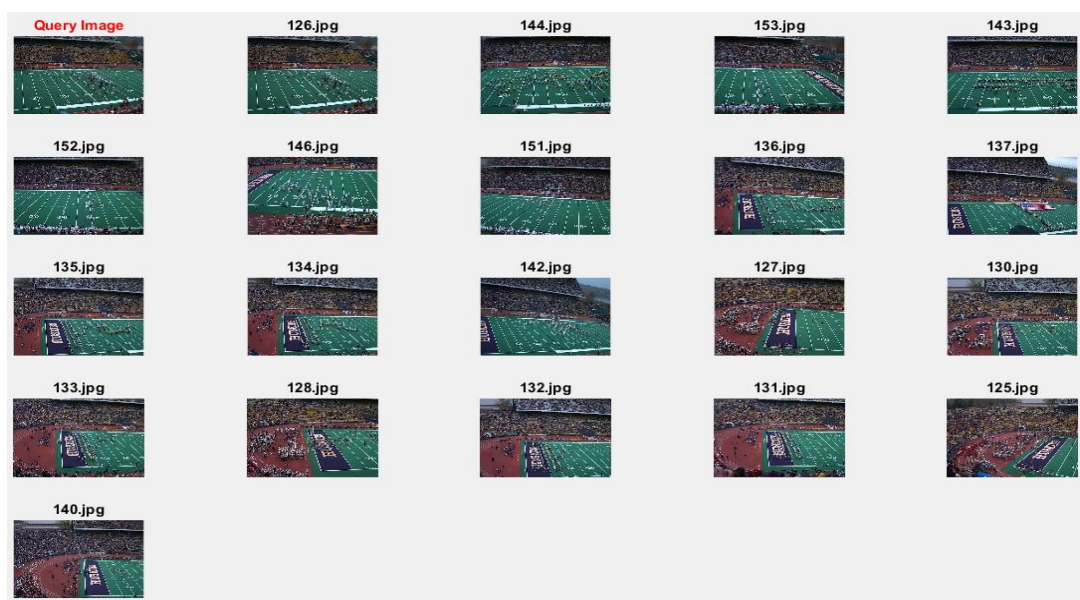

**Fig 12.** Image retrieval result for the class " Football".

The comparisons presented in **Table 5**, show that the proposed research outperforms the existing [8, 27, 30] in term of precision when using GT image dataset. Image retrieval results obtained using a query image from GT image dataset is represented in **Fig. 12.**

## 5. Conclusion and Future Directions

The research proposed in this paper is based on the feature fusion of low-level visual features that are based on color histogram, color moments and invariant moments. The proposed research in this paper reduces the semantic gap between image visuals concepts and feature vector representation. We have selected the combination of these low-level visual features as they are robust and computation of these features requires less time. The distance of the query image to the images placed in the dataset is computed by applying Manhattan distance. The Corel-1K, Corel-1.5K, Coil and GT image datasets are selected to compute the image retrieval performance of low-level feature fusion of color and shape features. The values of precision, recall and feature extraction obtained from the proposed low-level visual feature fusion are compared with existing research of CBIR. All of these comparisons reflect the higher performance of proposed research model. The proposed research is efficient in term of feature extraction. Keeping in view this efficiency, the proposed research can be applied in any real-time domain for computer vision such as image retrieval or video surveillance. In future, we aim to extend the proposed low-level feature fusion with a classification model such as deep auto encoders for large-scale image retrieval.

2013.